\documentclass{article}

\usepackage{PRIMEarxiv}

\usepackage[utf8]{inputenc} 
\usepackage[T1]{fontenc}    
\usepackage{hyperref}       
\usepackage{url}            
\usepackage{booktabs}       
\usepackage{amsfonts}       
\usepackage{nicefrac}       
\usepackage{microtype}      
\usepackage{lipsum}
\usepackage{fancyhdr}       
\usepackage{graphicx}       
\graphicspath{{media/}}     

\graphicspath{{figures/}}
\usepackage{xcolor}
\usepackage{bm}
\usepackage{enumerate}
\usepackage{booktabs}
\usepackage{siunitx}
\usepackage{algorithm}
\usepackage{algpseudocode}
\usepackage{amssymb}
\usepackage{amsmath}
\usepackage{multirow}

\renewcommand{\vec}[1]{\bm{\mathrm #1}}
\DeclareMathOperator*{\argmin}{argmin}
\newcommand{\method}[0]{eXIE}

\title{Explaining Image Enhancement Black-Box Methods through a Path Planning Based Algorithm
}

\author{
  Marco Cotogni, Claudio Cusano\\
  Dept. of Dept. of Electrical, Computer and Biomedical Engineering\\
  University of Pavia, Italy\\
  Via Ferrata 1, 27100, Pavia, Italy \\
  \texttt{marco.cotogni01@universitadipavia.it} \\
  \texttt{claudio.cusano@unipv.it} \\
}

\begin{document}
\maketitle

\begin{abstract}
Nowadays, image-to-image translation methods, are the state of the art for the enhancement of natural images.  Even if they usually show high performance in terms of accuracy, they often suffer from several limitations such as the generation of artifacts and the scalability to high resolutions.  Moreover, their main drawback is the completely black-box approach that does not allow to provide the final user with any insight about the enhancement processes applied. In this paper we present a path planning algorithm which provides a step-by-step explanation of the output produced by state of the art enhancement methods, overcoming black-box limitation. This algorithm, called eXIE, uses a variant of the A$^*$ algorithm to emulate the enhancement process of another method through the application of an equivalent sequence of enhancing operators. We applied \method{} to explain the output of several state-of-the-art models trained on the Five-K dataset, obtaining sequences of enhancing operators able to produce very similar results in terms of performance and overcoming the huge limitation of poor interpretability of the best performing algorithms.
\end{abstract}

\keywords{Explainable AI \and Image Enhancement \and Heuristic Search \and Path Planning}

In the last years deep learning drastically changed the fields of image processing, computer vision and computational photography.
For image enhancement, image-to-image models~\cite{isola2017image, gharbi2017deep, zhang2021star},  proved to be very effective solutions capable of achieving very high performance in terms of accuracy
In most image-to-image methods a low quality image is mapped via a neural network to its high quality version, preserving the details  and the semantic content.  Although these methods can be very effective, they come with important limitations: they typically work on images of a given fixed resolution, they can produce visible artifacts, and they do not provide easy to interpret results (black-box approach). This last limitation makes it difficult for a non-expert user to understand the enhancement steps applied by the algorithm. The lack of explainability makes it difficult their application in many fields (e.g. medical or legal) where the interpretability of the results is crucial.

The generation of artifacts in the output mainly depends on the additional challenging requirement for the underlying neural network to learn to preserve the content of the original image. 
Moreover, many image-to-image methods assume a fixed size for the input images to reduce the amount of memory and computation required. Therefore, high-resolution images, such as those  directly obtained from the sensor of DSLR cameras, are first downsampled to more manageable dimensions.  This is a reasonable choice when the resolution of the image is not critical but in fields where the resolution matters (e.g.\ computational photography), this is often unacceptable. 

%

%
%
%

\begin{figure}
        \centering
        \includegraphics[scale=0.44]{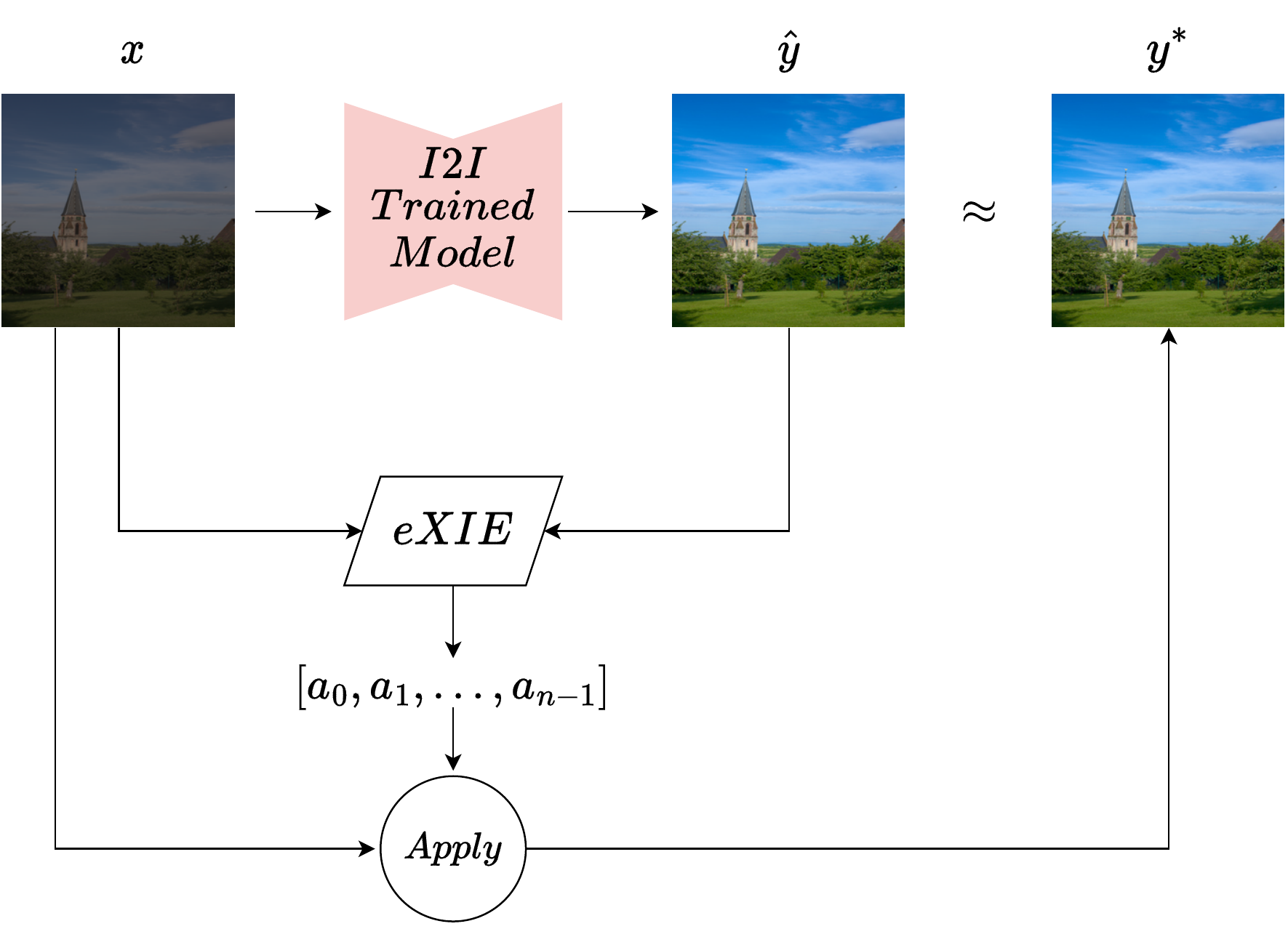}
        \caption{Schematic view of \method{}.  Given as input a low quality image $x$, a pretrained model for image enhancement is used to obtain an high quality version $\hat{y}$. Both, the low quality and the high quality versions of the images are used to execute a modified version of the A$^*$ algorithm in order to find the shortest sequence of enhancing operators $[a_0, a_1, \dots, a_{n-1}]$ that emulates the enhancement process. Once this sequence is obtained, each operator is applied to the low quality version to enhance it. }
        \label{fig:\method{}}
\end{figure}

In the light of these limitations, we present a new method, called \method{}, able to emulate the enhancement process of image-to-image translation models for image enhancement.
It is not a new independent enhancement method, but it is intended 
to provide explainable outputs for existing methods.
\method{} takes an input image together with its translated version obtained trough any other image enhancement method. It provides as output a sequence of enhancing operators that turns the input into an approximation of the output.  To do so, \method{} uses a modified version of the A$^*$ search algorithm.  Figure \ref{fig:\method{}} summarizes how the proposed method works. 

\method{} can be used in used in many application scenarios.  For instance it can be used to provide a baseline to professional photographers, that can revise and adjust the proposed processing pipeline instead of starting from scratch.  It can also be used as an educational tool to support beginners in understanding when to apply image processing operators.

To verify the quality of the sequences found by the proposed method we performed a thorough experimentation on the Five-K data set, comparing several other state of the art methods results with their corresponding version modified by \method{}. The loss in accuracy caused by \method{} was minimal, and in some cases the method was even able to improve the underlying enhancement method.  Moreover, \method{} showed impressive results when applied to high resolution images.

This paper presents the following main contributions:
\begin{itemize}
    \item The \method{} algorithm for the explanation of the output of existing image enhancement methods.  As far as we know, this is one of the first works combining path finding and image enhancement.
    
    \item \method{} is one of the first explainable algorithm appositely developed to obtain a step-by-step interpretation of image enhancement methods output predictions.
    
    \item  A novel heuristic function that is used by the modified A$^*$ algorithm.
    
    
    \item An in depth experimentation, in which we observed how \method{} was able to explain the outputs of several state of the art methods with minimal loss in performance as well as to reproduce with high fidelity human retouched images. 
    
    
\end{itemize}

The paper is organized as follows:
Section~\ref{sec:related} presents the most relevant works on image enhancement and explainable AI published in the literature. Then, Section~\ref{sec:method} describes the proposed method and how it works. Section~\ref{sec:results} presents the results obtained by applying the method on the images produced by several state of the art methods for image enhancement and by expert photographers. Finally, Section~\ref{sec:conclusions} concludes the paper with a discussion about possible directions for future investigation on this topic. 


\section{Related Works}
\label{sec:related}
Image enhancement is a classic problem in image processing in which a low quality image is transformed in its high quality version while preserving its visual content. Providing an explaination of the behavior of these methods, is very important to allow beginner photo editors to learn how to visually enhance low quality images using the output of these method as reference. In this section we analyze two different families of works related to our method: Image Enhancement methods and Explainable AI algorithms (XAI).

\subsection{Enhancement methods}
Modern approaches to image enhancement are typically based on deep convolutional neural networks.  In particular, in image-to-image translation methods a neural network learns how transform a low quality image into its enhanced version.  Isola et al.\ developed  a GAN-based conditional architecture able to translate the input image from a source domain to a new or a different target domain \cite{isola2017image}. Using a similar generative adversarial architecture, Zhu et al.\ developed a cycle-loss for translating images from an unpaired training set \cite{zhu2017unpaired}. Liu et al.\ presented an unsupervised version of the generative adversarial networks for image-to-image translation \cite{liu2017unsupervised}. Ronnenberg et al.\ developed an encoder-decoder architecture for image segmentation \cite{ronneberger2015u}. This architecture was adapted during the years for several tasks, including image enhancement. Cai et al.\ developed a modified version of the UNet architecture to enhance low light images \cite{cai2019low}.

Another family of methods includes the parametric approaches. In this case, the neural network learns the coefficients of a parametric color transformation to be applied to the low quality input image to obtain the enhanced version. Gharbi et al.\ developed an architecture able to learn the coefficients of a transformation working on a low resolution version of the original input image. Once the transformation is obtained, it is applied to the high resolution image to enhance it \cite{gharbi2017deep}. Bianco et al.\ proposed an architecture able to learn the parameters of a color transformation. These parameters are combined with a basis function and the resulting transformation is applied to the input image \cite{bianco2019learning}. The same research group also proposed another approach where a neural network estimates the coefficients of splines color curves~\cite{bianco2020personalized}. Song et al.\ presented an enhancement architecture based on a color curve encoder. This encoder computes the color curves parameters on a low resolution version of the input image. The learned transformation is then applied to the high resolution image \cite{song2021starenhancer}.

Recently, Zhang et al.\ presented a transformer based architecture for real time image enhancement \cite{zhang2021star}. In this work, a reduced structure of the original version presented by Vaswani et al.\ \cite{vaswani2017attention} is used to enhance low light images. Kim et al.\  \cite{kim2021representative} proposed representative color transformation to enhance low quality images. The proposed architecture is composed of an encoder, a feature fusion module, and two representative color transformation modules (local and global). Kim et al., presented a color representation learning method to enhance low-light images \cite{kim2022learning}. Finally, Hu et al.\ proposed a white-box RL algorithm able to enhance a low quality image providing the sequence of enhancing operators applied \cite{hu2018exposure}.

\subsection{Explainable AI}
In the last years, artificial intelligence (AI) and deep learning based methods showed super performances in several fields and applications as computer vision, natural language processing, time series analysis, etc.
One major drawback of AI methods is their intrinsic lack of interpretability.
Several works have been presented in order to provide explanations for neural networks' predictions, making them interpretable for the final user.

Grad-Cam is one of the most famous XAI algorithms. This method, following the gradient flow in a convolutional neural network, is able to provide a localisation map as output highliting the most relevant part of the image where the network was mainly focusing its attention to output the final prediction. Moreover, the method was also tested on different tasks such as visual question answering (VQA) and captioning proving the ability of the model of analyzing and providing an explanation of neural networks reasoning \cite{selvaraju2017grad}.

Similarly, Zintgraf et al.\ proposed a prediction difference analysis method to visualize the neural netowrks' predictions in the task of image classification. This method analyzes the prediction provided by the neural network and assigns a score to each input feature with respect to a chosen class. Unlike to Grad-Cam, this method, highlights part of the input feature map using conditional and multivariate sampling. This general approach was tested on two different domains: natual images from ImageNet dataset and medical magnetic resonance imaging scans \cite{Zintgraf2017vdn}.

Shirikumar et al.\ developed a method called Deep Learning Important FeaTures (DeepLIFT).This method, decompose the neural network's predictions on a defined input, backpropagating the contribution of each neuron of the network to the input features. A contribution score is assigned to each of the neuron in the CNN in an efficient way with only a single backpropagation step. This score is based on the difference between the activation of each neuron and a reference activation \cite{shrikumar2017learning}.

Lundberg et al.\ presented an XAI feature importance method based on SHapley Additive exPlanations values (SHAP). Given a prediction, a series of values are assigned to each feature accordingly with the importance of the singular feature in the classification process \cite{lundberg2017unified}.

Ribeiro et al, proposed a decision rules based method for explaining the behavior of complex model. The presented method, provides local sufficient conditions (highlightning parts of the input) for the correct prediction in several domains like text-classification, structured prediction, tabular classification, image classification and visual-question answering. \cite{ribeiro2018anchors}

One of the most powerful architectures for image generation are generative adversarial networks (GANs). This method shows high performance in several domains, producing great images with high quality details. One of the main drawback of this method is to be a complex black-box model. Nguyen et al.\ developed a GAN, based on activation maximization, to highlight the features learned by each neuron of the generator in order to explain and interpret the generation process\cite{nguyen2016synthesizing}.

Li et al.\ proposed an unrolling based method called DUBLID (Deep Unrolling for Blind Deblurring). The unrolling procedure was used to decompose an iterative algorithm, able to emulate in the gradient domain a total-variation regularization method, in to a fully interpretable neural network for image deblurring \cite{li2020efficient}.

Wang et al.\ developed an interpretable sparse coding network for image super resolution. The proposed architecture, composed of a cascade of spare-coding network with different scale factors, was able to obtain very good performances without the introduction of artifacts in the final images \cite{wang2015deep}.

\section{Method}
\label{sec:method}
The aim of the proposed method is to find an equivalent sequence of enhancing operators able to emulate the enhancement process of another state of the art method.  The search of the sequences is performed by a modified version of the A$^*$ algorithm \cite{hart1968formal}. A$^*$ is an efficient path finding algorithm, able to find the shortest path from an initial node to a final node in a graph, if it exists.

To find the shortest path between two nodes in a graph, the algorithm uses two functions to evaluate the nodes:
\begin{itemize}
    \item the backtrack function $g(x)$, that computes the length of the path from starting node to the current nod $x$e.
    \item The heuristic function $h(x)$ that estimates the length of the optimal path connecting the current node $x$ to the target node. In order to ensure the optimality of the solution, the heuristic must be optimistic, i.e.\ the estimates of the length of the path to the target node must not exceed the actual distance.
\end{itemize}
These two functions are added together and the resulting value $f(x)=g(x)+h(x)$ is assigned to each node. The algorithm iteratively visits one node at a time
following ascending $f$ values.  The search terminates when the final node is visited.


\begin{figure*}
    \centering
    \includegraphics[scale=0.58]{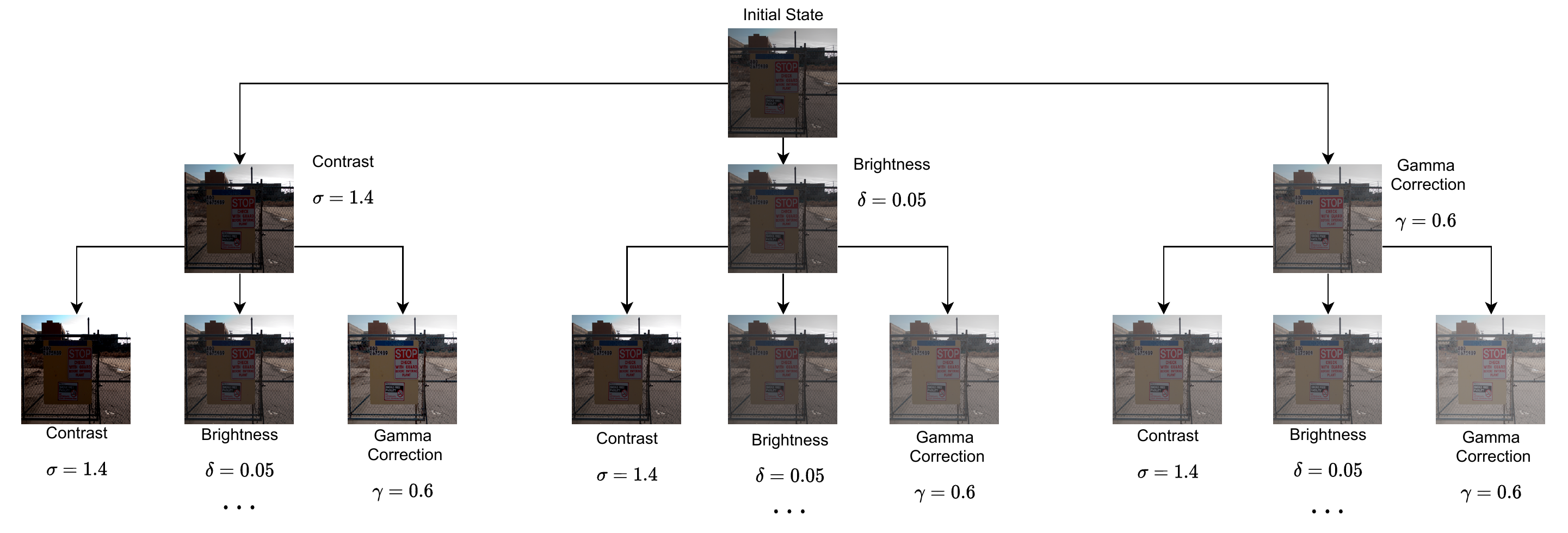}
    \caption{Example of the graph traversed by the \method{} algorithm. For space reasons, the considered graph is generated using only three editing operators over all the channels of the images and it is truncated after two levels. 
    }
    \label{fig:graph}
\end{figure*}

\subsection{\method{}}

In this work, the nodes composing the graph are images and the edges connecting two nodes are image processing operators. A connection between two nodes in the graph corresponds to a transition from an image (node) to its modified version obtained applying an editing operator to the former one.
Figure \ref{fig:graph} shows an example of the graph traversed by the search algorithm. 

The initial node is identified by the low quality image to be enhanced $x$. 
The goal is to find the shortest sequence of enhancing operators $[a_{0},a_{1},\dots,a_{n-1}]$ to improve the quality of the image $x$ ( formally, $y^*= a_{n-1}(a_{n-2}(\dots a_0(x)))$ as close as possible to the image produced by the considered image to image model $\hat y$.

As editing operators we considered a small set of general filtering functions widely used in image processing.  In the following $\vec{x}_{ijc}$ represents channel $c \in \{R, G, B\}$ of the pixel with $(i, j)$ coordinates:
\begin{itemize}
    \item \emph{Brightness adjustment}: 
    \begin{equation}
        \vec{x}_{ijc} \to \vec{x}_{ijc} + \delta.
    \end{equation}
    We considered the parameters
    $\delta \in \Delta = \{-0.05, +0.05, -0.005, +0.005\}$, and applied to all the color channels or to a single color channe.  
    \item \emph{Contrast adjustment}: 
    \begin{equation}
        \vec{x}_{ijc} \to \mu_c + \sigma \times (\vec{x}_{ijc} - \mu_c),
    \end{equation}
    where $\mu_c$ is the average channel value. We considered the variants with $\sigma \in \Sigma = \{0.9, 1.4\}$ and applied the operator channel wise considering all channels or just one.
    \item \emph{Gamma Correction}:
    \begin{equation}
        \vec{x}_{ijc} \to (\vec{x}_{ijc})^\gamma.
    \end{equation}
     We considered the two values $\gamma \in \Gamma = \{0.6, 1.05\}$, and applied the transformation channel wise considering all channels or just one.
\end{itemize}
Values and the operators contained in the three sets $\Delta$, $\Gamma$, $\Sigma$ have been selected in order to enhance the input image with a reasonable number of applications of the operators. Smaller values would lead to the same results but requiring a longer time for the searching process.
The value of the input pixels is always supposed to be in the $[0, 1]$ range, and all output values are clipped to stay in that range.
In total 32 image processing operators where considered.

\subsection{Heuristic Function}
The heuristic function is a key element in the search algorithm.
It estimates the length of the path from a given node to the target node. In order to guarantee the optimality of the solution it needs to be optimistic, that is, it needs to underestimate the actual distance between the nodes.  On the other hand, being too optimistic slows down the algorithm that would not be able to quickly progress towards the target. 

We defined the heuristic function by considering how many times one of the operators needs to be applied to transform a single pixel value into the target value.
More in detail, for each pixel value $\vec{x}_{ijc}$ we compute three counters: the Brightness Counter, the Contrast Counter and the Gamma Correction Counter. The Brightness counter is the number of times the brightness operator, needs to be applied to $\vec{x}_{ijc}$  to make it reach the target $\hat y_{ijc}$.
\begin{equation}
\label{eq:br}
    N^{(B)}_{ijc} = \min_{\delta \in \Delta} \frac{|\vec{x}_{ijc}- \hat y_{ijc}|}{|\delta|}.
\end{equation}

We defined the Contrast Counter in a similar way.  This requires to identify special cases since sometimes it is not possible to transform the pixel value in tothe desired target just by using this operator.
\begin{equation}
    \label{eq:co}
    N^{(C)}_{ijc} = 
    \begin{cases}
        \frac{1}{\log \max \Sigma} \log \frac{\hat y_{ijc}-\mu_c}{\vec{x}_{ijc}-\mu_c}  & \text{if } \vec{x}_{ijc}>\mu_c \text{ and } \hat y_{ijc}>\mu_c, \\
        \frac{1}{\log \min \Sigma} \log \frac{\hat y_{ijc}-\mu_c}{\vec{x}_{ijc}-\mu_c} & \text{if }  \vec{x}_{ijc}<\mu_c \text{ and } \hat y_{ijc}<\mu_c, \\
        \infty & \text{otherwise}.
    \end{cases}
\end{equation}

Finally, the Gamma Correction Counter is defined as:
%
\begin{equation}
    \label{eq:GA}
    N^{(G)}_{ijc} =
    \begin{cases}
        \frac{1}{\log \max \Gamma} \log \frac{\log \hat y_{ijc} }{\log \vec{x}_{ijc}}  & \text{if } \vec{x}_{ijc} \geq \hat y_{ijc}, \\
        \frac{1}{\log \min \Gamma} \log \frac{\log \hat y_{ijc} }{\log \vec{x}_{ijc}} & \text{if } \vec{x}_{ijc}< \hat y_{ijc}, \\
    \end{cases}    
\end{equation}
%
%
The heuristic function $h$ is defined for the whole image $\vec{x}$ as an upper bound to the number of applications of any given operator to match each pixel with the corresponding target value:
\begin{equation}
    \label{eq:h}
    h(\vec{x}) = \max_{ijc} \min_{a \in [B,C,G]}  N^{(a)}_{ijc}.
\end{equation}

Concerning the backtrack function $g(\vec{x})$ we counted the number of times an operator has been applied to the initial image to obtain the image $\vec{x}$.
In order to limit the searching time of the algorithm, we introduced two other modifications. The search terminates when the difference between the actual and the target nodes is below a set threshold $\|\vec{x} - \hat{\vec y} \| < \tau$.  Moreover, it can also terminate when the number of explored nodes exceeds a set limit $L$.  When this happens the visited node that is closest to the target is selected and the path (i.e. the sequence of enhancing operators) from the root to this node is obtained as output along with the enhanced image.  We observed that, when the number of visited nodes overcomes the value of $L=7000$, the accuracy of the output images tend to be very stable. Accordingly with the maximum number of explored nodes $L$ we set the value of the threshold $\tau=2$. These choices have been taken in order to deal with the trade-off between the output image quality and the time required to explore the graph.
The pseudo-code for the whole procedure is reported in Algorithm \ref{algo:a*}.

\begin{algorithm}
  \caption{\method{}}\label{algo:a*}
  \begin{algorithmic}
\State Input: image $\vec x$,\\
target $\hat{\vec y}$,\\
set of image operators $\mathcal{A}$,\\
target threshold $\tau$,\\
maximum number of visited nodes $L$.\\
\State Output: sequence of operators $\langle a_0, a_1, \dots a_{n-1} \rangle$ taken from $\mathcal{A}$, such that $a_{n-1}(a_{n-2}(\dots a_0(\vec x)))$ is approximately equal to $\hat{\vec y}$.    \\
\State $O \leftarrow \{\vec x\}$
\Comment{Open set}
\State $C \leftarrow \emptyset$
\Comment{Closed set}
\State $g(\vec x) \leftarrow 0$
\State $\text{path}(\vec x) \leftarrow \langle \rangle$
\While {$|C| < L$}
    \State $\vec x_\text{current} \leftarrow \arg\min_{\vec x' \in O} g(\vec x') + h(\vec x')$
    \State $C \leftarrow C \cup \{\vec x_\text{current}\}$
    \State $O \leftarrow O \setminus \{\vec x_\text{current}\}$
    \If {$\| \vec x_\text{current} - \hat{\vec y} \| < \tau$} 
        \State \textbf{return} $\text{path}(\vec x_\text{current})$
    \EndIf
    \For{$a \in \mathcal{A}$}
        \State $\vec x_\text{new} \leftarrow a(\vec x_\text{current})$
        \State $g(\vec x_\text{new}) \leftarrow g(\vec x_\text{current}) + 1$
        \State Compute $h(\vec x_\text{new})$ according to (\ref{eq:h})
        \State $\text{path}(\vec x_\text{new}) \leftarrow \text{path}(\vec x_\text{current}) \oplus \langle a \rangle$
        \Comment{append $a$}
        \State $O \leftarrow O \cup \{\vec x_\text{new}\}$
    \EndFor
\EndWhile
\State $\vec x_\text{best} \leftarrow \argmin_{\vec x' \in C \cup O} \| \vec x' - \hat{\vec y} \|$
\State \textbf{return} $\text{path}(\vec x_\text{best})$
\end{algorithmic}
\end{algorithm}


\section{Results}
\label{sec:results}
In the first part of this section we present the dataset we used in our experiments. Then we discuss the results obtained by applying \method{} to several image enhancement methods from the state of the art.  Finally, we will assess the performance of \method{} obtained with a new baseline neural network especially designed for this purpose.

\subsection{Dataset}
The dataset used for our experiments is the Adobe-MIT Five-K dataset \cite{fivek}. This dataset is composed of 5000 high-resolution images in RAW format. For each of these images, five enhanced version are included, each one enhanced by an expert photographer identified by a letter from A to E. In our experiments we used the RAW images and the images retouched by Expert C which is considered the most consistent of the five. We split the images by following the procedure presented by Hu et al.~\cite{hu2018exposure}: \num{4000} training images and \num{1000} test images.

\subsection{Enhancement methods}
We selected seven popular state-of-the-art methods for image enhancement. They belong to different families of approaches such as image-to-image translation , parametric, reinforcement learning and transformer based methods:
\begin{enumerate}
    \item Exposure: This deep reinforcement learning based method, is able to enhance low quality images producing high quality images~\cite{hu2018exposure}.
    \item CycleGan: GAN based method, uses a cycle loss for learning the correct function to map an input image from a source domain to a target domain. Among all the possible applications, it could be also applied to enhancement tasks. It could suffer from the artifacts generation in output images~\cite{zhu2017unpaired}.
    \item DaR: This method, called Distort and Recover, is a double Q-learning based algorithm for image enhancement~\cite{park2018distort}. 
    \item Pix2Pix: Proposed by Isola et al.\ in 2016, this conditional adversarial network for image-to-image translation showed great performances in several domains of application. \cite{isola2017image}.
    \item HDRNet: This architecture based on bilateral grid processing and local affine color transformation, learns the correct transformations to be applied to the low quality high resolution input image observing its resized version~\cite{gharbi2017deep}.
    \item Star-DCE: transformer based method for image enhancement. This method, splits the input images in patches and embed them into tokens. Then these tokens are passed to a long-short range Transformer module composed of two branches: one for long-range context (composed of a cascade of transformer blocks) and the other for short-range context (composed of a cascade of convolutions and batch normalizations~\cite{zhang2021star}.
    \item Parametric: Proposed by Bianco et al, this pipeline learns, in a paired training scenario, the parameters of a color transformation, using a downsampled version of the high resolution low quality input image. Once the color transformation is obtained, it is applied to the original input image to enhance its content~\cite{bianco2019learning}.
\end{enumerate}
All these methods have been trained using the following configuration described in the original paper on the \num{4000} images from the Five-K dataset. The trained models were then evaluated on the \num{1000} test images.
Finally, all the images obtained have been analyzed by \method{} to produce the corresponding sequences of operators. 

\subsection{Experimental results}
We used four different metrics: the Peak Signal-to-Noise Ratio (PSNR) \cite{kelecs2021computation}, Learned Perceptual Image Patch Similarity (LPIPS) \cite{zhang2018unreasonable}, Delta E ($\Delta E$) \cite{goodman2012colour} and Structural Similarity Index SSIM\cite{hore2010image}. Table~\ref{tab:results} compares the metrics computed on the output images of the considered methods and those computed on the images produced by \method{}.

\begin{table}[t]
\caption{Comparison of state-of-the-art enhancement methods on the Adobe Five-K dataset with and without \method{}.}
\begin{center}
\begin{tabular}{llcc}
\toprule
Metric &   Method     & Original & \method{} \\
\midrule
LPIPS$\downarrow$ & Exposure~\cite{hu2018exposure}&
\num{0.16}  & 0.13  \\
& CycleGan~\cite{zhu2017unpaired}    &  \num{0.16} &  0.14 \\
& DaR~\cite{park2018distort}    & \num{0.10}  &  0.11 \\
& Pix2Pix~\cite{isola2017image}    & \num{0.09}  & 0.12  \\
& HDRNet~\cite{gharbi2017deep}    & \num{0.08}  &  0.10 \\
& Star-DCE~\cite{zhang2021star}    & \num{0.08}  &  0.09 \\
& Parametric~\cite{bianco2019learning}    & 0.07  &  0.10 \\
\addlinespace
PSNR$\uparrow$ & Exposure~\cite{hu2018exposure}& \num{18.74}  & 19.45  \\
& CycleGan~\cite{zhu2017unpaired}    & \num{19.38}  & 19.38  \\
& DaR~\cite{park2018distort}    & \num{20.91}  & 21.00  \\
& Pix2Pix~\cite{isola2017image}    & \num{23.05}  & 21.41  \\
& HDRNet~\cite{gharbi2017deep}    & \num{22.31}  & 21.55  \\
& Star-DCE~\cite{zhang2021star}    & 23.55  &  22.41 \\
& Parametric~\cite{bianco2019learning}    & \num{23.20}  & 22.26  \\
\addlinespace
$\Delta E\downarrow$ & Exposure~\cite{hu2018exposure}& \num{13.57}  &13.06 \\
& CycleGan~\cite{zhu2017unpaired}    &  \num{12.95} & 13.88  \\
& DaR~\cite{park2018distort}    & \num{13.05}  &  11.09 \\
& Pix2Pix~\cite{isola2017image}    & \num{8.84}  & 11.03  \\
& HDRNet~\cite{gharbi2017deep}    & \num{9.53}  & 10.66  \\
& Star-DCE~\cite{zhang2021star}    & 8.45  & 9.75  \\
& Parametric~\cite{bianco2019learning}    &  \num{8.52} & 9.86  \\
\addlinespace
SSIM$\uparrow$ & Exposure~\cite{hu2018exposure} & \num{0.81} & 0.83   \\
& CycleGan~\cite{zhu2017unpaired}    & \num{0.78}  &  0.82 \\
& DaR~\cite{park2018distort}    &  \num{0.86} & 0.86  \\
& Pix2Pix~\cite{isola2017image}    & \num{0.86}  & 0.86  \\
& HDRNet~\cite{gharbi2017deep}    & \num{0.89}  &  0.87 \\
& Star-DCE~\cite{zhang2021star}    & \num{0.89}  &  0.88 \\
& Parametric~\cite{bianco2019learning}    & 0.90  &  0.88 \\
\bottomrule
\end{tabular}
\end{center}
\label{tab:results}
\end{table}

In all the cases \method{} was able to imitate the underlying enhancement method with a good level of approximation.  The lost in terms of the four metrics considered is never very high, and in some cases the images found by \method{} are even better than the starting ones.  This happens because by design \method{} prevents the introduction of artifacts.  In practice, artifacts are harder to reproduce than correct enhancement.
Only for the most accurate methods, and only for the PSNR and $\Delta E$ metrics, the difference is noticeable.  This drop in accuracy is acceptable in many applications, if the explainability of the model can be ensured.
Differences in terms of LPIPS and SSIM are always negligible.

\begin{figure*}
    \centering
    \includegraphics[scale=0.63]{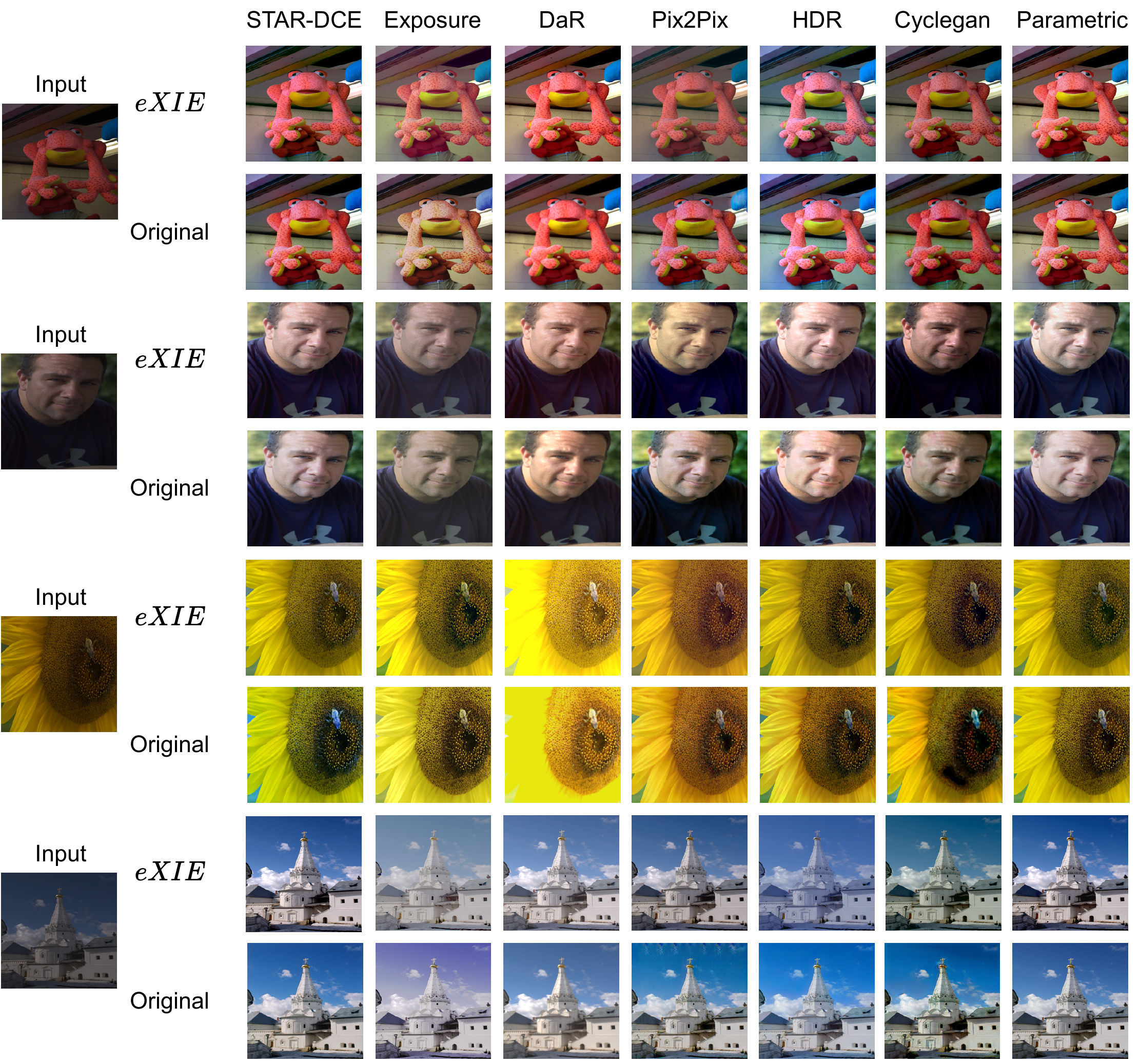}
    \caption{Comparison of the images obtained by \method{} and by the original methods.}
    \label{fig:comparison}
\end{figure*}

From the qualitative analysis of the images produced by \method{} shown in Figure~\ref{fig:comparison} it is possible to notice how \method{} is able to emulate the enhancement process of the original methods very accurately.  For instance, by looking at the images produced by \method{} on CycleGan (column 6) it is possible to see that the artifacts have been removed and the color balance for the images obtained by \method{} is better than the version produced using the image-to-image translation method.  Moreover, by observing the results on DaR (column 3) and Exposure (column 2), we can see that \method{} is able to obtain better saturation and contrast in the final image with respect to the result of the original models.
By observing the results on the best models instead, Star-DCE (column 1), Pix2Pix (column4), HDR (column 5) and Parametric (column 7), it is clear how \method{} is able to mimic them with high fidelity. 

\subsection{Sequence inspection}
\begin{figure}
    \centering
    \includegraphics[scale=0.57]{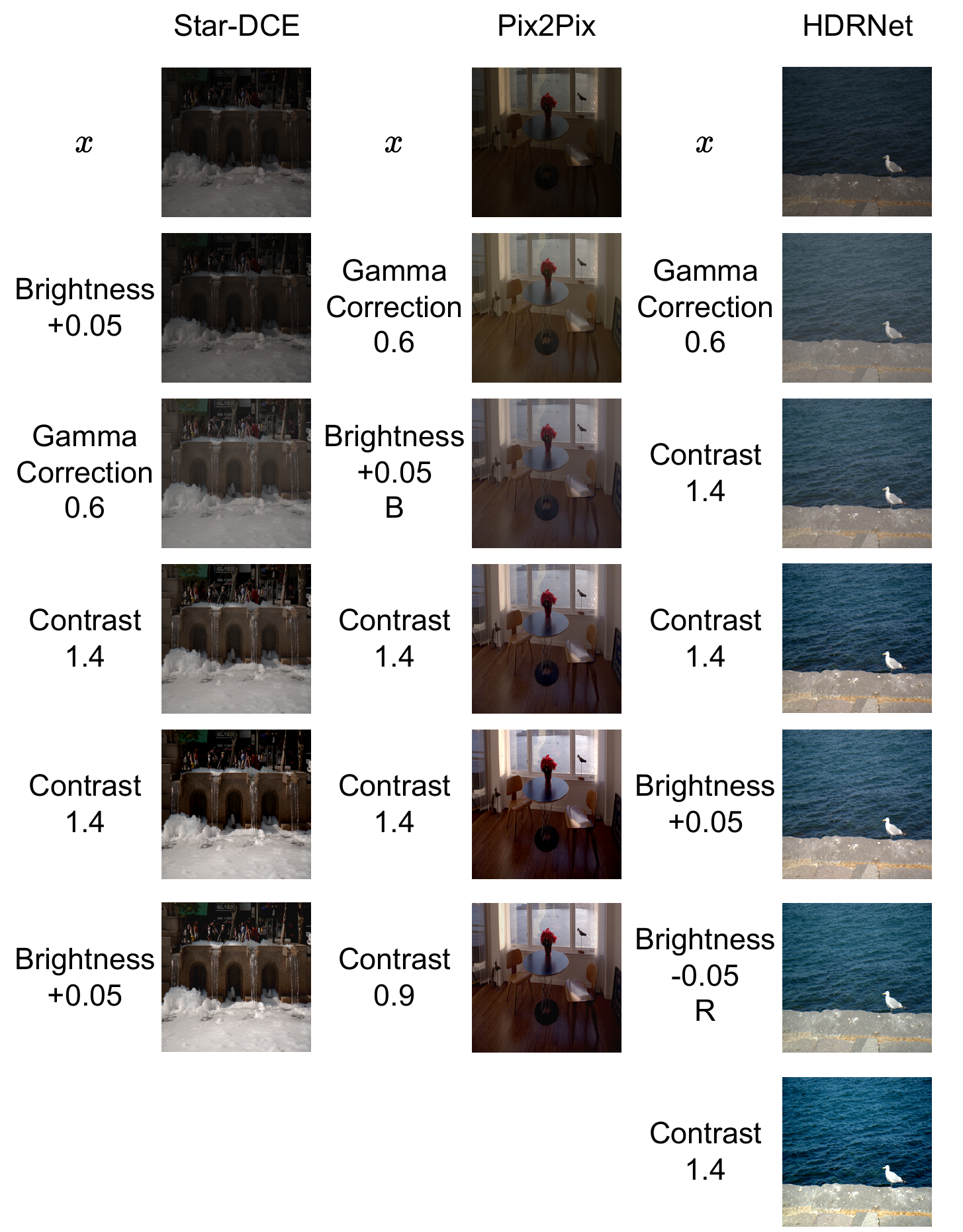}
    \caption{Example of the sequences obtained with the application of \method{} on Star-DCE, Pix2Pix and HDRNet.}
    \label{fig:sequence}
\end{figure}

As a further analysis we studied the sequences produced by \method{} (Figure~\ref{fig:sequence}). The sequences obtained with the considered state of the art methods are quite heterogeneous among them. Moreover, they follow a different order in the application of the operators.

The firsts operators applied are typically those that increase the global brightness of the image.  These operators, are applied on all the pixels of the three channels of the image. Then, when the image reaches a good overall balancing, more fine-grained operators are applied. For instance, in the HDRNet column the figure shows how the algorithm finds that, after the application of the brightness over the all image, the best next action is to decrease the value of the red channel of the image, obtaining a refinement of the color distribution.

\subsection{Enhancement of high resolution images via UNET}

Many image-to-image-translation methods, like pix2pix or cyclegan,  work with images of fixed dimensions. For these methods, the enhancement of very high resolution images without producing artifacts, requires to keep constant the spatial dimension of the features is kept constant along all the layers of the model.  We addressed this scenario by using our method on very low-resolution versions of the input images and by applying the sequences found to the original high resolution images.  This approach, requires very few computational resources in terms of memory and time and, as we will see, it does not decrease output quality.

\begin{figure}
    \centering
    \includegraphics[scale=0.53]{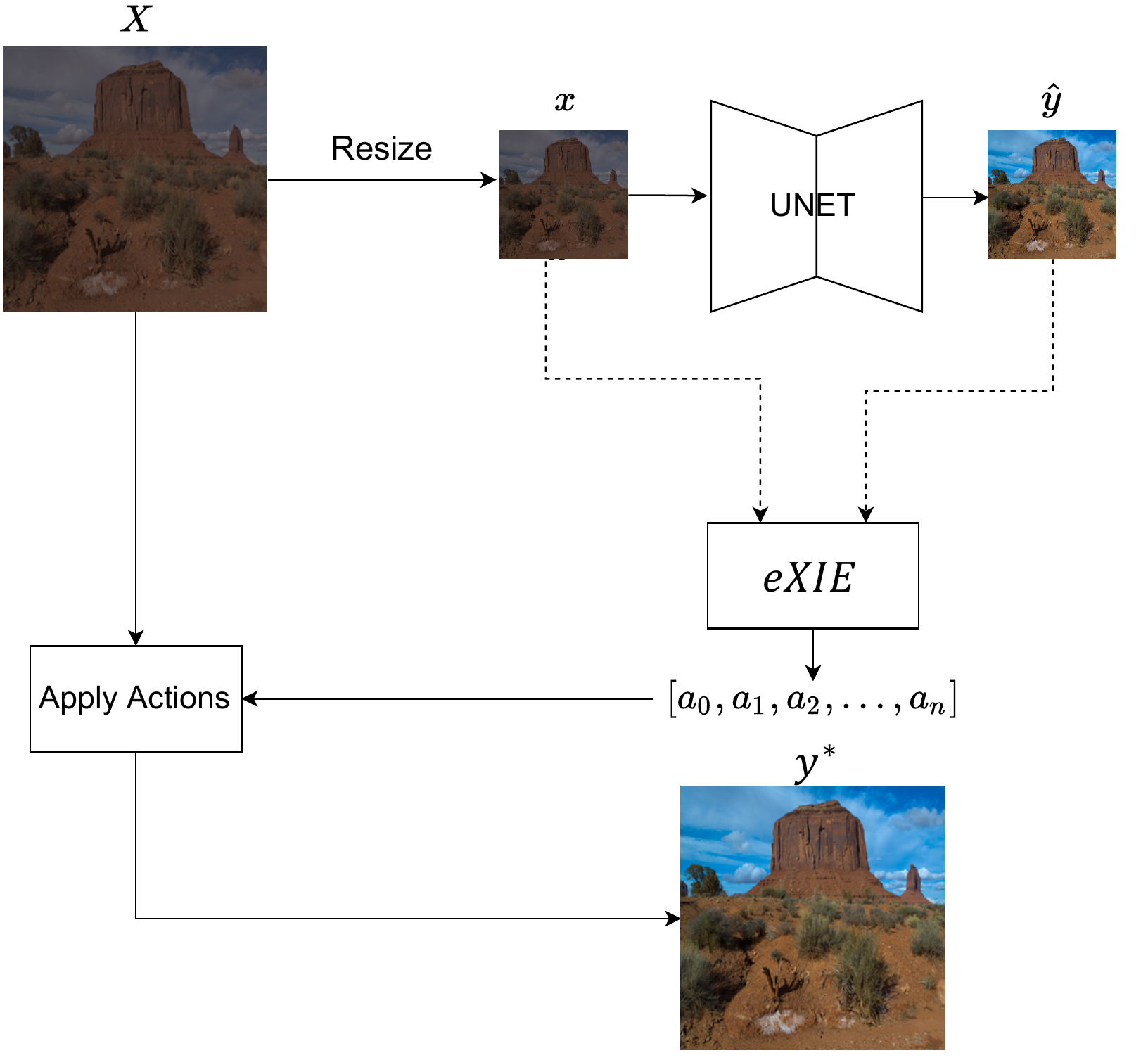}
    \caption{The high resolution input image $X$, is resized obtaining a low resolution version $x$. This image is enhanced using the previously trained UNet. This architecture provide as output the image $\hat y$. The images $x$ and $\hat y$ are used to execute the \method{} algorithm and obtaining the sequence of enhancing operators to enhance the high resolution image $X$ and obtaining $\hat Y$.}
    \label{fig:unet}
\end{figure}

More in details, given a high-resolution image from the Five-K dataset, we resized it to a very low-resolution ($32 \times 32$) and provided as input to an especially designed convolutional neural network based on the UNet~\cite{ronneberger2015u} architecture.  \method{} is then applied to the resulting low-resolution images, and the sequence of operators found is applied to the high-resolution input image. Figure \ref{fig:unet} summarizes this approach. 

The neural network used in this experiment is composed of two parts: an encoder and a decoder. Each of the four  blocks in the encoder includes a convolutional layer with kernel of size $4 \times 4$ and stride 2, batch normalization and leaky relu as activation function.
The decoder is composed of four blocks, each of them including a transposed convolutional layer (of kernel size $4 \times 4$ and stride 2), batch normalization, relu and dropout ($p=0.05$). The output layer of the architecture has a sigmoid function as activation function in order to restrict the values of output pixels to the range $[0, 1]$.

During the training of the network, the high resolution input images $X$ from the Five-K training set are resized obtaining a low resolution version $x$. These images are given as input to the net obtaining the enhanced version $\hat y = f(x;\theta)$. The images $\hat y$ are compared with the target images $y$ using the binary cross entropy (BCE) as loss function:
\begin{equation}
    BCE = - \frac{1}{HWC} \sum_{c=0}^{C}\sum_{i=0}^{W}\sum_{j=0}^{H} y_{jic} \cdot \log(\hat y_{jic}) + (1-y_{jic})\cdot \log(1-\hat y_{jic})
\end{equation}

The model has been trained using standard data augmentation techniques (cropping, resizing, random flip and rotations) on the image pairs for 600 epochs and batch size 32. The UNet parameters are updated using mini batch gradient descent and AdamW as optimizer \cite{loshchilov2017decoupled} with a learning rate of 5e-3. The learning rate was decayed of 0.1 every 100 epochs starting from the 200th.

Once the network has been trained, it has been used to enhance the low-resolution versions of the images from the Five-K test set. These enhanced images have been then processed by \method{} along with the original low-resolution inputs.

\begin{table}[t]
    \caption{Results of the experiment with the neural network for low-resolution enhancement (on low-resolution images) and the application of \method{} (on high-resolution images).}
    \begin{center}
        \begin{tabular}{lcc}
        \toprule
         Metric & UNet on Low-Res & \method{} on High-Res\\
        \midrule
        PSNR$\uparrow$&23.36&22.09\\
        $\Delta E\downarrow$ &8.55&10.31\\
        SSIM$\uparrow$&0.93&0.88\\
        \bottomrule
        \end{tabular}
        \end{center}
    \label{tab:unet}
\end{table}

Table~\ref{tab:unet} summarizes the results of the application of \method{} to the low-resolution images. The first column shows the performance of the neural model on the low resolution images. The metrics reported in the second column, are computed on the high resolution images. \method{} shows good performance even when it is applied to very high resolution images as the original ones provided in the Five-K dataset.

The values of the metrics computed on the images enhanced using \method{} are good.  In fact, by comparing these results with those showed in Table~\ref{tab:results} we can see that \method{} is able to enhance high resolution images better than other methods like Exposure, CycleGan and Dar.  The values of the performance metrics are very similar to those obtained by HDRNet.

\begin{figure}
    \centering
    \includegraphics[scale=0.45]{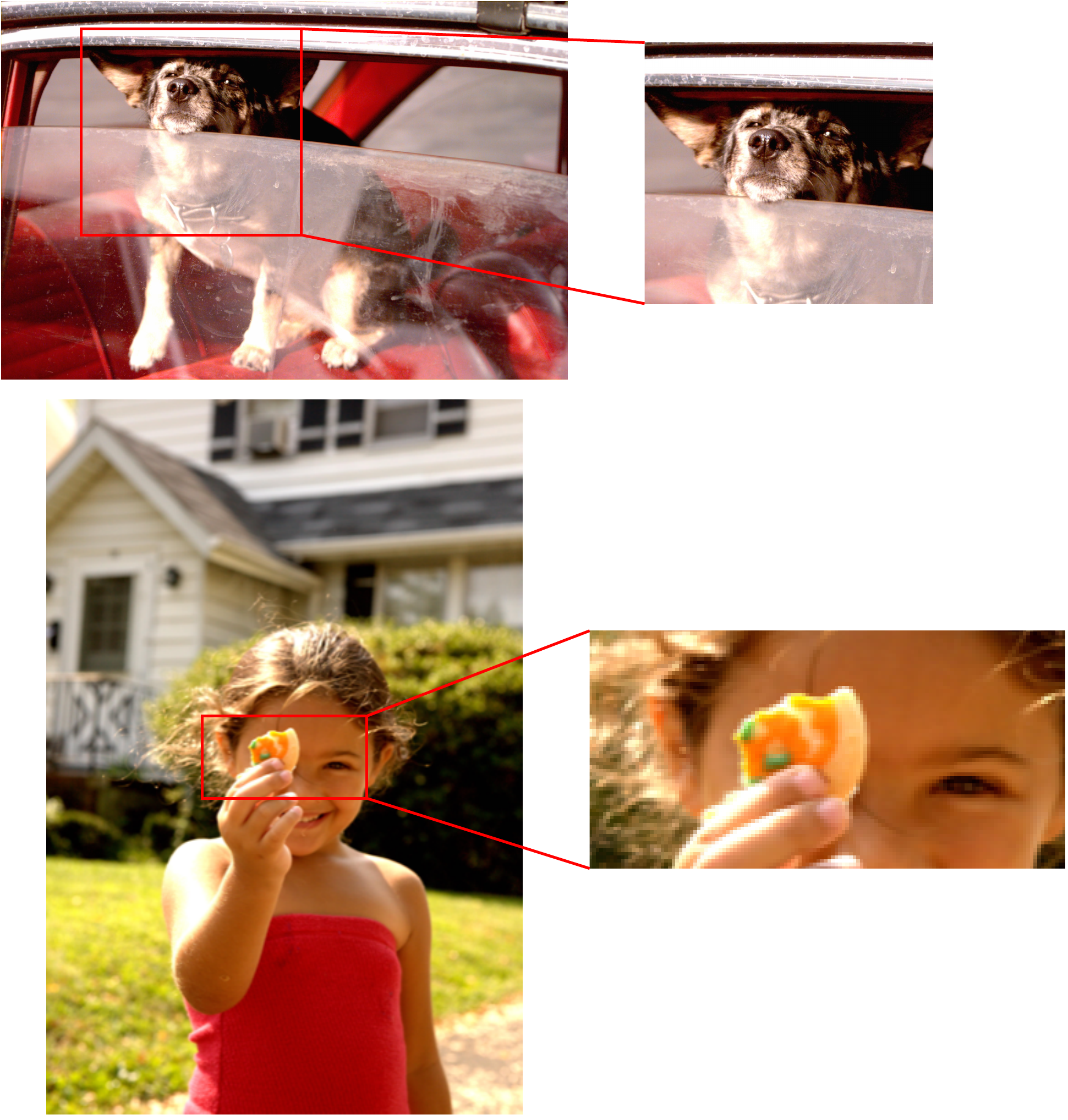}
    \caption{Example of the application of \method{} on two high resolution image from Five-K test set after being processed by the UNet.}
    \label{fig:highres}
\end{figure}

Analyzing the images showed in Figure \ref{fig:highres}, is it possible to notice the abscence of artifacts and that the color balancing applied by \method{} is correct . Moreover, from the analysis of the details is it possible to see that the visual content of the image is preserved correctly (this is also confirmed by the high value of SSIM in the previous table).

\subsection{Case Study: Human Target}
As a last experiment, we used \method{} to reverse engineering the work of an expert photo editor.  The goal, here, is to replicate the image enhanced by a human expert as a sequence of elementary editing operations.  The application for this scenario is educational: a beginner photo editor could use the system to learn how to achieve a given editing effect by looking at how \method{} reduced it to a sequence of operations. 

To do so, we applied \method{} on the images in the Five-K dataset with the goal of reproducing the versions enhanced by the experts. Table~\ref{tab:experts} summarizes the results obtained.
\begin{table}[]
    \centering
    \caption{Results of the application of \method{} on the test images of the Five-K dataset using the images of the 5 available experts as target.}
    \begin{tabular}{lcccc}
    \toprule
    Metric & LPIPS$\downarrow$ & PSNR$\uparrow$ & $\Delta E\downarrow$ & SSIM$\uparrow$\\
    \midrule
    expA & 0.01 & 27.76 & 5.83 & 0.92\\
    expB & 0.01 & 29.14 & 5.16 & 0.95\\
    expC & 0.02 & 25.44 & 7.43 & 0.91\\
    expD & 0.01 & 28.02 & 5.73 & 0.94\\
    expE & 0.02 & 26.50 & 6.69 & 0.93\\
    \bottomrule
    \end{tabular}
    \label{tab:experts}
\end{table}

The results in the Table \ref{tab:experts} shows that \method{} is able to provide very high quality images reproducing the enhancement process applied by the experts of the Five-K dataset. Moreover, the values showed for the considered metrics are very high confirming the ability of our algorithm to emulate not only the enhancement process of image-to-image translation models, but also the sequence of operations chosen by human experts

\begin{figure*}
    \centering
    \includegraphics[scale=0.35]{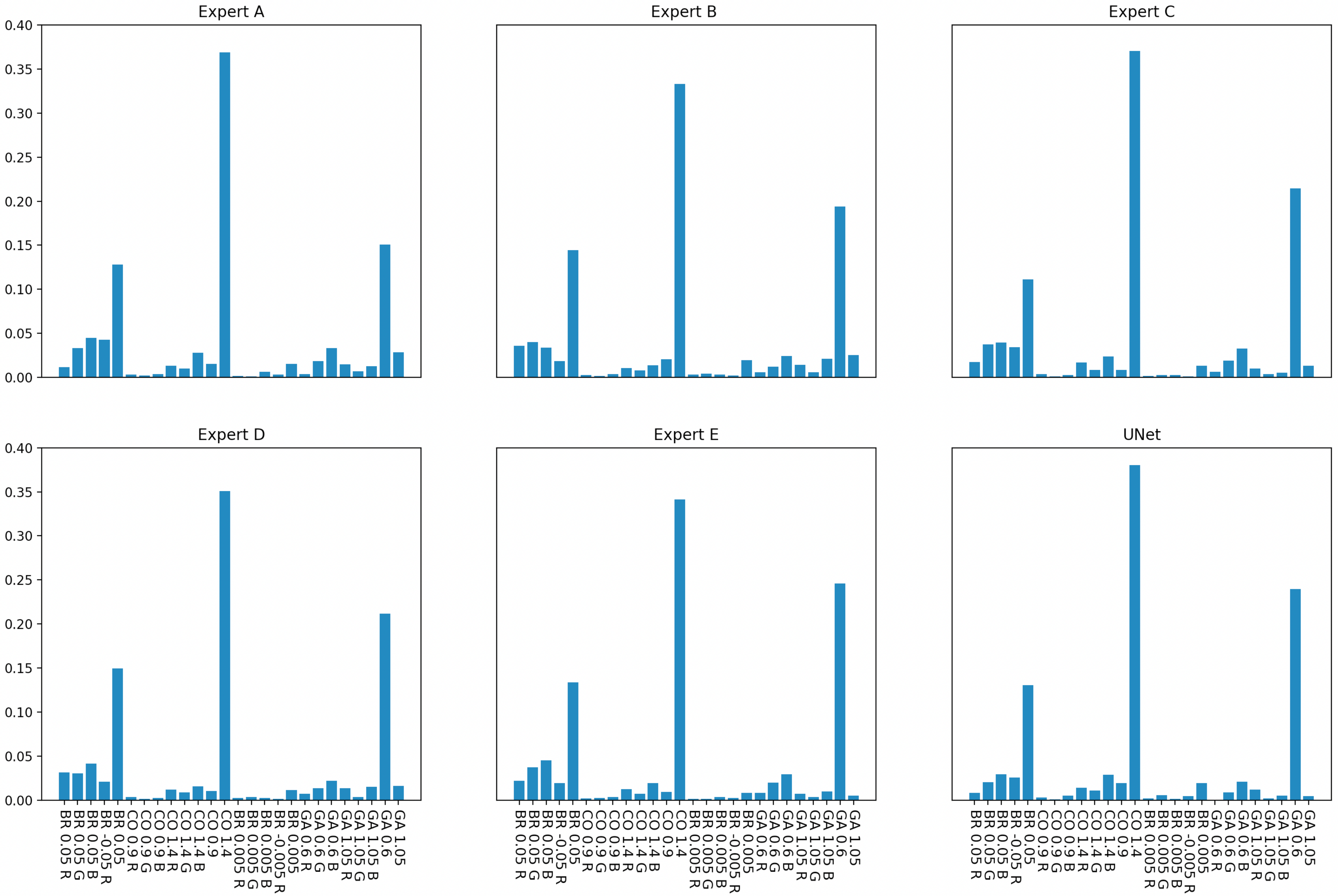}
    \caption{Actions distributions of the sequences obtained applying \method{} to replicate the enhancing process applied by Five-K experts and the UNet.}
    \label{fig:actions}
\end{figure*}

From the analysis of the distributions of the operations forming the sequences selected by \method{} (Figure \ref{fig:actions}), it is possible to notice that the most frequently selected operations are those that modify all the color channels. The distributions are quite similar for all the experts, and also for the UNet, with just some difference in the frequency of single-channel operations (for the brightness, in particular)

Observing actions probability distributions of each single expert it is possible to notice general trends or particular preferences. By observing Expert A actions distribution for example, it is possible to notice how brightness and gamma correction over all the three channels have almost the same probability, indicating that these two actions are interchangeable for Expert A most of the times.

\section{Conclusions}
\label{sec:conclusions}
In this paper we proposed \method{}, a novel method for explaining the enhancement process of state-of-the-art methods for image enhancement. This XAI method is able to provide an equivalent sequence of enhancement operators that emulate the behavior of image enhancement methods with only a small loss in the performances. 

\method{}, was able to produce good looking images, even with a better color distribution with respect to some of the methods on which it is applied. Moreover, its output images do not present artifacts and show high quality details.

The ability of generalization of the method has been tested by applying it to high resolution images and executing the heuristic search algorithm on their low resolution versions. 

In the future we plan to explore the application of \method{} to different processing tasks like retargeting and restoration.  We will also consider more specific domain of applications such as that of medical imaging, where the explainability of the enhancement process is of vital importance.
\bibliographystyle{unsrt}  
\bibliography{eXIE}

\end{document}